\documentclass[11pt,a4paper]{article}
\usepackage{times,latexsym}
\usepackage[hyphens]{url}
\usepackage[T1]{fontenc}
\usepackage{booktabs}
\usepackage{stfloats}
\usepackage{multirow}
\usepackage{graphicx}
\usepackage{amsmath}
\usepackage[export]{adjustbox}
\usepackage{tabularray}
\usepackage{tipa}
\usepackage{pgfplots}
\pgfplotsset{compat=1.17} % Ensures compatibility with a stable version
\usepackage{hyperref}

\usepackage{microtype}

\usepackage{xspace,mfirstuc,tabulary}

\usepackage[acceptedWithA]{tacl2021v1} 
\newif\iftaclinstructions
\taclinstructionsfalse % AUTHORS: do NOT set this to true
\iftaclinstructions
\renewcommand{\confidential}{}
\renewcommand{\anonsubtext}{(No author info supplied here, for consistency with
TACL-submission anonymization requirements)}
\newcommand{\instr}
\fi

\iftaclpubformat % this "if" is set by the choice of options

\else

\fi

\usepackage[multiple]{footmisc}
\def\sectionautorefname~#1\null{\S#1\null}
\def\subsectionautorefname~#1\null{\S#1\null}
\def\subsubsectionautorefname~#1\null{\S#1\null}

\title{The Impact of Automatic Speech Transcription on Speaker Attribution}

\author{
\centering
\hskip-1.0cm
\begin{tabular}{c@{\hskip .7in}c}
  Cristina Aggazzotti & Matthew Wiesner
  \\
  \textnormal{Johns Hopkins University} & \textnormal{Johns Hopkins University}
  \\
 \textnormal{\texttt{caggazz1@jhu.edu}} & \textnormal{\texttt{wiesner@jhu.edu}}
 \\
 & \\
  Elizabeth Allyn Smith & Nicholas Andrews
  \\
  \textnormal{Universit\'e du Qu\'ebec \`a Montr\'eal} & \textnormal{Johns Hopkins University}
  \\
 \textnormal{\texttt{smith.eallyn@uqam.ca}} & \textnormal{\texttt{noa@jhu.edu}}
   \end{tabular}
}

\date{}

\begin{document}
\maketitle
\begin{abstract}
Speaker attribution from speech transcripts is the task of identifying a speaker from the transcript of their speech based on patterns in their language use. This task is especially useful when the audio is unavailable (e.g. deleted) or unreliable (e.g.\ anonymized speech). Prior work in this area has primarily focused on the feasibility of attributing speakers using transcripts produced by human annotators. However, in real-world settings, one often only has more errorful transcripts produced by automatic speech recognition (ASR) systems. In this paper, we conduct what is, to our knowledge, the first comprehensive study of the impact of automatic transcription on speaker attribution performance. In particular, we study the extent to which speaker attribution performance degrades in the face of transcription errors, as well as how properties of the ASR system impact attribution. We find that attribution is surprisingly resilient to word-level transcription errors and that the objective of recovering the true transcript is minimally correlated with attribution performance. Overall, our findings suggest that speaker attribution on more errorful transcripts produced by ASR is as good, if not better, than attribution based on human-transcribed data, possibly because ASR transcription errors can capture speaker-specific features revealing of speaker identity.
\end{abstract}

\section{Introduction}\label{sec:intro}Authorship attribution is the task of identifying an author based on what they write (content) and how they write it (style) using various textual features. When the text comes from speech (via transcription) instead of writing, we refer to it as \emph{speaker attribution}\footnote{Note that we are not referring to speaker-attributed ASR~\citep{fiscus2006, kanda2020}, which aims to determine ``who spoke what'' (i.e.~using ASR and diarization) and focuses on correctly attributing words to who spoke them rather than identifying who the speakers are.} from speech transcripts. This task is closely related to speaker recognition, which focuses more on the acoustic and phonetic properties of the audio to identify or verify a speaker by their voice. Speaker recognition models generally rely on neural speaker embeddings produced from speech, such as x-vectors~\cite{snyder2018} and wav2vec 2.0 embeddings~\cite{baevski2020}, which can encode a range of paralinguistic (e.g.~emotion~\cite{cai2024, ulgen2024}) and linguistic (e.g.~language and accent~\cite{foley2024, pratap2024}, lexical content~\citep{raj2019}) information. However, these models are not specifically trained to encode these linguistic features, cannot operate directly on textual representations of speech, and cannot benefit from training on much larger text corpora that also encode many of these characteristics. Authorship attribution models, however, do rely on such textual features, and previous speaker attribution work has applied textual authorship attribution methods to speech transcripts, concluding that, at least in certain conditions, there is domain transfer of these models from text to speech~\citep{aggazzotti2024, hansen2023}.

Furthermore, although speaker recognition systems can be quite accurate in certain conditions (an equal error rate as low as 1\% or less \citep{desplanques2020, ravanelli2021, zeinali2019}, there are several instances in which the audio might be unavailable or unreliable, and thus attributing a speaker based on a transcript of their speech is the only viable means of identification. For example, with more accurate and accessible tools for voice cloning (duplicating someone's voice) and voice conversion (changing the style of someone's voice while keeping the content the same), a voice might not be a reliable indicator of identity but the content and style of what is said can be used instead~\cite{fang2019,yoo2020,panariello2024}. Also, video or audio files may not be kept due to storage or privacy restrictions with only the transcript preserved. Even when both are available, applying speaker recognition and speaker attribution to identify the same person via different features (voice and speaking patterns) provides more confidence in the identification. 

One drawback of having to automatically produce a transcript is the introduction of errors resulting from transcribing speech into text. With enough time and resources, speech can sometimes be transcribed manually, producing more accurate and faithful transcripts to the audio. However, for large quantities of data and for faster results, transcription usually occurs automatically via an automatic speech recognition (ASR) system, which naturally produces more transcription errors. Despite a recent increased use of ASR systems for transcription, no previous work has examined how these errors impact speaker attribution performance. 

We thus aim to determine how more errorful automatic transcription affects speaker attribution performance. To do this, we run a variety of ASR systems on the same audio to obtain transcripts with a range of error rates (\autoref{exp:wer}). Then, we test multiple authorship attribution models out-of-the-box on each automatic transcript to see how transcription error rates affect speaker attribution performance (\autoref{exp:auto}). The results are surprising: word error rate does not seem to significantly impact attribution performance. We explore various possible reasons for why error rate does not seem to correlate with attribution performance and propose that many of the errors made by the ASR systems capture speaker idiosyncrasies and thus are revealing of speaker style (\autoref{exp:limit}). To find the floor in performance, we test transcripts with error rates over 90\% and discover that, when content is no longer reliable or no longer available, models avail themselves of speaker utterance length, so performance does not degrade significantly (\autoref{exp:highwer}). However, utterance length may be an artifact of the dataset used, so although it may be useful for this dataset, it might not be a reliable feature in other settings. Next, we train a probing classifier for a variety of features on the training set verification trials but find that performance only improves for models that incorporate the content of what is said (\autoref{exp:ft}). Finally, in many real-world settings, training and evaluation data may differ, creating a train-test mismatch.~\autoref{exp:mismatch} explores one such setting: attribution models trained on gold standard transcripts do not generalize well to ASR transcripts. 
\section{Related Work}\label{sec:related}\citet{doddington2001} was the first to look at lexical features of speech, namely unigrams and bigrams, rather than exclusively short-term (i.e.~often from a single frame), acoustic features for speaker recognition. Using human transcriptions of the Switchboard corpus (part of the expanded version of the 2001 NIST Speaker Recognition Evaluation~\citep{nist2001}),~\citet{doddington2001} found that frequently-occurring bigrams perform surprisingly well at single speaker detection. 

At the time, speaker recognition systems were struggling to identify the same speaker across different channels and varying amounts of noise; therefore, copying from clues humans use to identify a speaker, studies began considering ``high-level'' features (i.e.~linguistic features extracted from longer time spans than a single frame), such as word usage and frequencies, to improve speaker recognition methods. Methods relying on lexical patterns~\citep{canseco2004}, phone n-grams~\citep{andrews2002, lei2007}, conditional pronunciation modeling~\citep{klusacek2003, leung2004}, prosody~\citep{adami2003, shriberg2005}, and combinations of high-level and low-level features~\citep{weber2002, campbell2003, kajarekar2003} became popular, especially due to better ASR systems and increased data availability. However, these methods were more computationally expensive and required longer audio samples compared to their acoustic method counterparts. Therefore, when the latter improved with technological advances like deep learning (e.g.~i-vectors~\citep{dehak2011}, x-vectors~\citep{snyder2018}), explicitly incorporating high-level features became less common in mainstream uses; nevertheless, speaker vectors were found to encode some of these features to an extent, such as speaking rate, lexical information, and utterance length~\cite{raj2019} as well as speaker demographic attributes~\citep{luu2021,wu2024}.  

In forensic use cases, though, where interpretable features and high accuracy are paramount, experts continued to explicitly examine some high-level features, such as idiosyncratic word use, in combination with acoustic and auditory features~\citep{shriberg2008, foulkes2012, sergidou2024}. Recently, the more comprehensive examination of lexical features of speech has resurfaced with the application of textual authorship analysis techniques to transcripts of speech. For instance, some forensic work has used frequent-word analysis, or analyzing the frequencies of function words, to distinguish speakers in human-transcribed datasets~\citep{scheijen2020, sergidou2023, sergidou2024}. In addition, the PAN competition on authorship verification, which historically included only written data, briefly expanded to include both written and spoken (human-transcribed interviews and image descriptions) data~\citep{pan2023}. Concurrently,~\citet{hansen2023} created a large dataset of speech transcripts containing mostly automatically transcribed human speech as well as LLM generations of simulated spoken texts. Testing various authorship models without incorporating topic control, they found that n-gram-based and transformer-based models performed well on human spoken texts, but not as well on machine-generated data. Other work has leveraged pretrained language models fine-tuned on speech transcripts of each speaker for text-based speaker classification, and then combined these results with those of audio-based speaker recognition systems for improved speaker identification on data with explicit name introductions or well-known speakers~\citep{zamana2024}. 

\citet{aggazzotti2024} also explored the transferability of text-based attribution models to the spoken domain by testing various neural and non-neural textual authorship attribution models on verification trials of human-transcribed conversational speech transcripts. They found that textual attribution models did transfer to speech transcripts unless the topic discussed by the speakers was controlled for, in which case attribution performance across all models decreased, sometimes very significantly. Their work focused on human-transcribed speech, or gold standard transcripts, thus creating a potential ceiling benchmark for authorship attribution of speech transcripts, although we find that even the human transcripts are not always entirely faithful to the audio. The present work creates a similar benchmark, but for automatically transcribed speech with varying amounts of errors.~\citet{campbell2003} briefly tested a word n-gram speaker recognition system on various automatic transcripts, finding that performance was fairly robust even on transcripts with a 50\% word error rate; our results on a more comprehensive range of experiments support this finding.

\section{Data and Methods}\label{sec:methods}The \citet{aggazzotti2024} speaker attribution benchmark\footnote{Available at \href{https://github.com/caggazzotti/speech-attribution}{github.com/caggazzotti/speech-attribution}.} uses the Fisher English Training Speech Transcripts corpus~\citep{fisher}, a dataset of telephone conversations between strangers in which participants were given a specific topic to discuss and participate in multiple calls. These factors as well as providing audio and two styles of human transcription, being large for a manually transcribed dataset (11,699 $\sim$10-minute calls over 1,960 hours), spanning a range of specific topics, and gender balancing the speakers make this dataset uniquely suited for our study. 

The authors split the data into training (50\%), validation (25\%), and test (25\%) sets with different speakers in the training set and evaluation sets. Each of these splits was composed of three difficulties based on the level of conversational topic control, each containing roughly the same number of same speaker (`positive') and different speaker (`negative') trials. The difficulty levels were based on the level of conversational topic control: `base' had no topic control, `hard' restricted the speaker in positive trials to discuss different topics and the speakers in negative trials to discuss the same topic, and `harder' used the same positive trials as `hard' but the negative trials contained two speakers in the same call, thus not only talking about the same topic, but also the same subtopics throughout the conversation. Since similar trends were found across all three difficulties, we focus on the `hard' setting (959 positive trials, 985 negative trials, 1474 speakers) in the main text and include the `base' results in~\autoref{sec:appendix} for comparison. 

\begin{table*}[t]
\centering
\resizebox{\linewidth}{!}{%
\normalsize
\begin{tabular}{@{\hskip 2pt}c@{\hskip 4pt}|c|l}
\bf cpWER & \bf ASR & \bf Transcription \\
\toprule
n/a & Authors & yeah th- th- that's that's what i i extremely agree that it's uh it's uh just like the soap operas \\ 
``0\%'' & Gold & yeah th- that's that's what i i extremely agree that it's uh it's just like that soap operas \\ 
14\% & Giga & yeah this this one i extremely agree that it's ah it's ah it's just like the soap opera \\
19\% & Assembly & yeah that's what i extremely agree that uh it's just like the soap opera \\
21\% & Swb & yeah th- th- this is this is what i- i i extremely i agree that it's a it's a it's just like the soap all barage\\ 
26\% & Whisper & uh yet that that that's what i think extremely agree that uh it's uh just like that hope operas \\ 
32\% & TED & yeah this i extremely agree that it 's just like the soap operas \\ 
\bottomrule
\end{tabular}
}
\caption{\small Transcriptions of the same utterance by the authors, Fisher (gold standard), and various ASR systems with a range of cpWERs (calculated on all calls in the test set). Even Fisher is not 100\% faithful to the audio. In this example, the ASRs mainly differ in how they transcribe filler words, restarts, and non-standard pronunciations.}
\label{tab:transcr}
\vspace{-9pt}
\end{table*}

\subsection{ASR systems}\label{sec:asr}
We choose five ASR systems that produce transcriptions across a range of accuracies, presented here in order of increasing error rate.~\autoref{tab:transcr} gives an example transcription and the error rate of each of these ASR systems but will be discussed in more detail in~\autoref{exp:wer}. Three are open-source ASR systems (available via HuggingFace), for which we know the data on which the systems were trained and how they work; the other two are commercial systems, AssemblyAI and Whisper, so we cannot know for sure which data they saw during training and there is a chance they could have been trained on the Fisher data, for instance.\footnote{According to personal communication with a head AssemblyAI researcher, Fisher most likely did not appear in the AssemblyAI training data.}

The first system is a zipformer transducer model that was trained strictly on \textsc{Gigaspeech},\footnote{\href{https://huggingface.co/yfyeung/icefall-asr-gigaspeech-zipformer-2023-10-17}{huggingface.co/yfyeung/icefall-asr-gigaspeech-zipformer-2023-10-17}} a corpus of 10,000 hours of high quality audio from audiobooks, podcasts, and YouTube across a range of topics. The second is Universal-1 by \textsc{AssemblyAI}, which was trained on 12.5 million hours of multilingual (English, Spanish, French, German) audio and purportedly has a lower word error rate (WER) than other systems, such as Whisper (large-v3).\footnote{\href{https://www.assemblyai.com/blog/announcing-universal-1-speech-recognition-model}{assemblyai.com/blog/announcing-universal-1-speech-recognition-model}} We keep all default settings except the following: we include disfluencies, exclude punctuation, and use dual channels. The third is a wav2vec 2.0 with connectionist temporal classification model that was fine-tuned on the \textsc{Switchboard} corpus.\footnote{\href{https://huggingface.co/speechbrain/asr-wav2vec2-switchboard}{huggingface.co/speechbrain/asr-wav2vec2-switchboard}} 

Next is the other commercial system: OpenAI's \textsc{Whisper} is a transformer sequence-to-sequence model that was trained on 680,000 hours of diverse and multilingual labeled audio.\footnote{\href{https://github.com/openai/whisper}{github.com/openai/whisper}} Specifically we use the `turbo' model, which is a faster and optimized version of `large-v3' with minimal accuracy degradation. The final system has the same architecture as the Gigaspeech system but was trained on \textsc{TED-LIUM3},\footnote{\href{https://huggingface.co/desh2608/icefall-asr-tedlium3-zipformer}{huggingface.co/desh2608/icefall-asr-tedlium3-zipformer}} which is 452 hours of audio from TED talks. Because of the formatting of its training data, this system adds a space before all apostrophes in contractions (e.g.\ \emph{can 't}) and transcribes `??' for unknown segments. Since we would like a range of transcription styles and errors, we did not remove these features.

\subsection{ASR transcription comparison}\label{exp:wer}  
The Fisher corpus includes both audio files and the corresponding manually transcribed transcripts, which have ``gold standard'' time segments for each speaker's utterances. Therefore, to obtain ASR transcriptions of the Fisher trials, we first collect the audio files of all calls present in the test set trials across all three difficulty levels. Because the ASR systems do not have built-in diarization, for each call, we extract the gold standard time segments for each turn, slice the audio file into separate turns, and then transcribe each speaker's turns using each ASR system. The resulting transcripts per speaker are then recombined to form a full transcript for that call. This is done for each call across all ASR systems except AssemblyAI, which transcribes each speaker separately by recording channel.

To measure the quality of automatically transcribed transcripts, we calculate their WER across all test set calls. WER is a widely-used measurement of the number of errors present in a transcript compared to the reference transcript. In our case, the reference transcript is the human-transcribed Fisher speech transcript corpus. To help focus on the errors we are most interested in---transcription errors rather than transcription style differences---and since some of the ASR systems do not transcribe capitalization and have limited punctuation, we use the LDC encoding of the Fisher corpus instead of the BBN encoding. The BBN encoding has capitalization and grammatical punctuation marks, such as periods, question marks, and commas, and notably segments utterances by who speaks (i.e. as soon as Speaker B says something, even if it is only a backchannel, Speaker A's turn ends). In contrast, the LDC encoding has text in all lowercase, has limited punctuation, namely hyphens and apostrophes, and groups utterances together despite interrupting speech from the other speaker. In addition, it includes non-speech sounds in brackets (e.g.~`[noise]', `[laughter]') and the annotator's hypothesized annotations of unclear speech in double parentheses (e.g.~`(( oh okay ))'). However, many ASR systems do not include these, so for the WER comparison only, we remove all bracketed non-speech sounds including the text inside, and we remove double parentheses while keeping their text. We similarly normalize any ASR transcripts that include brackets around non-speech sounds (only for WER), capitalization, or grammatical punctuation marks (but keep the TED-LIUM3 apostrophe space and double question mark). 

To avoid errors resulting from different segmentations across ASR systems, we use the concatenated minimum-permutation WER (cpWER) rather than the standard WER.\footnote{The standard WER is slightly higher than the cpWER for each ASR system but consistently so across all systems, so we choose the cpWER as a presumably more specific measure of transcription errors excluding errors resulting from segmentation differences.} The cpWER concatenates all speaker utterances in the reference transcript and the hypothesis transcript then computes the WER between the reference and all possible speaker permutations of the hypothesis and chooses the one with the lowest WER~\citep{chime6}. Concatenating all utterances per speaker ensures that differences in utterance segmentation do not count as errors. As an example, say one system segments Speaker A's utterances into two because Speaker B interrupted with a backchannel (e.g.~\emph{hmm}), but another system combines Speaker A's utterances together and includes Speaker B's backchannel after. As long as all the words are the same between the two versions, cpWER would not count the difference in utterance boundary as an error, whereas standard WER would. Thus, cpWER is a fairer measure across ASR systems that potentially segment differently. 

We calculate the cpWER using the MeetEval toolkit~\citep{meeteval23}\footnote{\href{https://github.com/fgnt/meeteval}{github.com/fgnt/meeteval}} across all test set transcribed audio files, which includes all calls present in all difficulty levels even if only one of the speakers appeared in a trial. We do this for each ASR system and compare to the normalized, manually transcribed Fisher LDC test data. \autoref{tab:transcr} gives a comparison of the differences in transcriptions and the kinds of errors made across ASR systems for an example utterance. The first row is the authors' manual transcription of the audio, which differs only slightly from Fisher (row 2), the gold standard reference transcription, in capturing an additional restart \emph{th-} and filler word `uh', and correctly transcribing \emph{the soap operas} instead of \emph{that soap operas}. The authors found that the gold standard transcriptions were not 100\% faithful to the audio, but errors were minor and fairly rare. The subsequent rows show the ASR systems in increasing order of error rate. The main differences lie in how and whether the ASR systems transcribe restarts and repeated words and how they transcribe the final word, which the speaker pronounces as /\textipa{"6.p@."\*r@z}/ (AH.puh.RUHZ) rather than the more common /\textipa{"6.p\*r@z}/ (AH.pruhz). %/ˈɒ.pə.'ɹəz/, /ˈɒ.pɹəz/

\subsection{Authorship attribution models}
We use five authorship attribution models, including the top four performers from~\citet{aggazzotti2024} for sake of comparison.\footnote{Note that all of these models were trained on written text, but the previous work found them to transfer reasonably well to speech transcripts (without strong topic control).} The first is Sentence-BERT (\textsc{SBERT}),\footnote{\href{https://huggingface.co/sentence-transformers/all-MiniLM-L12-v2}{huggingface.co/sentence-transformers/all-MiniLM-L12-v2}} which uses semantic similarity so is known to focus on the content of the text~\citep{sbert}. As a more stylistic counterpart to SBERT, Content-Independent Style Representations (\textsc{CISR}),\footnote{\href{https://huggingface.co/AnnaWegmann/Style-Embedding}{huggingface.co/AnnaWegmann/Style-Embedding}} focuses more on writing style than content by controlling the topic of verification trials at training time~\citep{wegmann2022}. The third model is Learning Universal Authorship Representations (\textsc{LUAR}),\footnote{\href{https://huggingface.co/rrivera1849/LUAR-MUD}{huggingface.co/rrivera1849/LUAR-MUD}} which makes use of more invariant stylistic features and often performs well in cross-domain settings~\citep{rivera-soto2021}. To more closely resemble the conversational speech transcripts, we choose a LUAR model that was trained on a large conversational dataset of Reddit comments by one million authors~\citep{khan2021}. The fourth model is \textsc{PANgrams}, which uses TF-IDF-weighted character 4-grams and is the PAN 2023 competition authorship verification baseline~\citep{pan2023}.\footnote{\href{https://github.com/pan-webis-de/pan-code/tree/master/clef23/authorship-verification}{github.com/pan-webis-de/pan-code/tree/master/clef23/authorship-verification}} Beyond these top four performers from previous work, as another content-independent style-based model, we also include the recent StyleDistance (\textsc{StyleD}),\footnote{\href{https://huggingface.co/StyleDistance/styledistance}{huggingface.co/StyleDistance/styledistance}} which uses verification trials of synthetically generated paraphrases that more closely control for content and style at training time \citep{patel2025}.

\section{Experiments}\label{sec:experiments}We first focus on the performance of the attribution models out-of-the-box since the amount of data for training is limited. We test the out-of-the-box attribution models on each ASR transcript (\autoref{exp:auto}) and then examine various limitations of using WER as a metric for predicting attribution performance (\autoref{exp:limit}). We also push the error rate very high and perform different ablations to find floor performance on this dataset (\autoref{exp:highwer}). Then we train a probing classifier on different feature representations of the training verification trials in an attempt to improve performance (\autoref{exp:ft}) and finally test how generalizable models that have been trained on gold standard data are on ASR transcripts (\autoref{exp:mismatch}). 

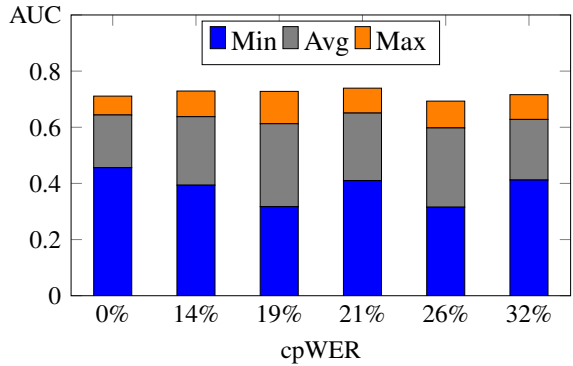
\begin{figure}[t]
\begin{center}
\hspace*{-.6cm} 
\begin{tikzpicture}[scale=0.95] 
    \begin{axis}[
        ybar stacked,
        ymin=0, ymax=1,
        symbolic x coords={0\%, 14\%, 19\%, 21\%, 26\%, 32\%},
        xtick=data,
        bar width=15pt,
        legend style={at={(0.5,0.98)}, anchor=north},
        legend columns=-1, % Puts all legend items in a single row
        xlabel={cpWER}, % X-axis label
        xlabel style={font=\small},
        xticklabel style={font=\small},
        ylabel style={at={(axis description cs:0,1.03)}, anchor=south, rotate=-90}, % Moves y-label to the top ticklabel
        ytick={0, 0.2, 0.4, 0.6, 0.8, 1.0}, % Define y-ticks
        yticklabels={0, 0.2, 0.4, 0.6, 0.8, AUC}, % Replace 1 with "AUC"
        yticklabel style={font=\small},
        height=5.5cm,  % Set a specific height for the plot
        width=8.5cm  % Keep the width the same
    ]
        % Min values
        \addplot [fill=blue] coordinates {(0\%,0.456162) (14\%,0.393698) (19\%,0.317226) (21\%,0.410023) (26\%,0.315309) (32\%,0.412115)};
        % Difference between min and avg
        \addplot [fill=gray] coordinates {(0\%,0.187597) (14\%,0.243980) (19\%,0.295371) (21\%,0.240568) (26\%,0.282341) (32\%,0.215809)};
        % Difference between avg and max
        \addplot [fill=orange] coordinates {(0\%,0.067215) (14\%,0.091372) (19\%,0.115194) (21\%,0.088683) (26\%,0.095491) (32\%,0.088188)};
        
        \legend{Min, Avg, Max}
    \end{axis}
\end{tikzpicture}
\vspace{-.5cm}
\caption{\small Minimum, average, and maximum (each color's height) attribution model AUC performance (y-axis) on `hard' difficulty level gold standard Fisher test verification trials (0\%) and several automatic transcript test trials with various cpWERs (x-axis). Despite significant increases in cpWER, attribution performance surprisingly stays fairly constant.}\label{fig:minavgmax}
\end{center}\vspace{-13pt}
\end{figure}

\subsection{Attribution on ASR transcripts}
\label{exp:auto}
We obtain an out-of-the-box speaker representation of the ASR-transcribed utterances of each speaker in a verification trial. We first take each ASR system's transcriptions of the calls and put them back into the benchmark verification trials by difficulty level. Next, we run each attribution model on each speaker's utterances per trial to obtain embeddings of their transcribed speech. Then, we calculate the cosine similarity between each representation in a trial and evaluate these similarities using the area under the ROC curve (AUC) to determine whether the speakers are the same or different. We do this procedure for each ASR system's transcriptions across all three difficulty levels; however, the trend is fairly consistent across difficulty levels on all experiments throughout the paper, so we choose to focus on the intermediate `hard' setting (but~\autoref{tab:fullboxbase} in~\autoref{sec:appendix} has the `base' level results for comparison).

\paragraph{Attribution performance does not correlate with WER.}
\autoref{fig:minavgmax} gives AUC attribution performance of the five attribution models for each transcription version of the test verification trials based on cpWER, ranging from the human-transcribed Fisher LDC transcripts at (presumably close to) 0\% cpWER to TED-LIUM3 at 32\% cpWER. The height of each color on the bar indicates the minimum, average, and maximum value. Higher AUC scores are better, and chance is 0.5. Surprisingly, overall attribution performance across automatic transcripts is similar to, and sometimes better than, performance on the Fisher gold standard transcripts, even at fairly high cpWERs---the increased errors do not seem to negatively impact the models' ability to distinguish speakers. 

\begin{table}
\scalebox{0.82}{
\centering
\hspace*{-.2cm}
\begin{tabular}{@{\hskip 2pt}r@{\hskip 4pt}|cccccc} 
\multirow{2}{*}{\bf AUC $\uparrow$} & \bf Gold & \bf Giga & \bf AAI & \bf Swb & \bf Whspr & \bf TED \\
& \bf 0\% & \bf 14\% & \bf 19\% & \bf 21\% & \bf 26\% & \bf 32\% \\
\toprule
\bf SBERT & \bf0.456 & 0.394 & 0.317 & 0.410 & 0.315 & \underline{0.412} \\
\midrule
\bf StyleD & 0.678 & \underline{0.700} & 0.671 & \bf 0.717 & 0.693 & 0.686 \\
\midrule
\bf LUAR & 0.711 & \underline{0.729} & 0.728 & \bf0.739 & 0.669 & 0.716 \\
\bottomrule
\end{tabular}
}
\caption{\small AUC performance of a content (SBERT; worst), style (StyleD; 2nd best), and content$+$style (LUAR; best) model on test verification trials with various cpWERs (columns). Best performance per model is bolded, second best underlined. All differences within each model are significant ($p < 0.01$; paired t-test). SBERT stays below chance; StyleD, LUAR generally perform better on ASR transcripts than on the gold standard. }
\label{tab:luarbox}
\vspace{-10pt}
\end{table}
 
Drilling down into these overall results to see if any models are particularly influenced by different ASR systems, we find that some of the models \emph{improve} on the ASR transcripts compared to on the gold standard. We show the three most representative models in~\hyperref[tab:luarbox]{Table 2}, but all models' performance is in~\autoref{tab:fullbox} in~\autoref{sec:appendix}. SBERT already performs below chance (0.5) on the gold transcripts but performs even worse on the ASR transcripts. In contrast, StyleD and LUAR perform higher than on the gold standard, except on AssemblyAI and Whisper, respectively. This improved performance suggests that the ASR errors might in fact be helpful for distinguishing speakers. 

\subsection{Limitations of WER as a metric for attribution}\label{exp:limit} 
WER strictly counts substitutions, deletions, and insertions, so any minor differences in the transcription, even those that do not impact attribution performance, are counted. For instance, if the Fisher transcription uses \emph{umm}, but an ASR system transcribes \emph{um}, the missing second \emph{m} would count as an error. Such irrelevant differences would inflate the WER for that ASR system without presumably impacting attribution performance. However, some ASR errors might be beneficial. For example, if a speaker has non-standard pronunciation, is sick, or is mumbling, the ASR system might transcribe a word phonetically as the speaker says it, while the human would transcribe the actual word. Therefore, the ASR system would have a higher WER but might capture some speech features that are more revealing of speaker style and identity. In such cases, an ASR transcription is actually \emph{better} than the human transcription for attribution.\footnote{The ideal would be to use an ASR system that is maximally faithful to the audio to capture each speaker's idiosyncratic features. We tried using CrisperWhisper~\citep{wagner2024}, a verbatim version of Whisper that strives to be faithful to the audio, but found that it does not consistently transcribe more faithfully than the gold standard and it produced similar results to Whisper. Future work should explore ways of improving CrisperWhisper or alternative methods of achieving verbatim transcription.} Work in other areas has also found similar limitations to WER, such as higher ASR WER leading to better speech translations~\citep{he2011} and WER not necessarily being a good indicator of speech understanding accuracy~\citep{esteve2003,wang2003}. In an attempt to explore how much differing transcriptions contribute to WER but do not negatively impact attribution performance, we compare the WER when content words are masked and when so-called style words, such as filler words, are masked.

\paragraph{More errors occur on style than content words.}\label{exp:masked}
For a rough estimate of how many of the errors contributing to the cpWER are so-called style (or function) words versus content words, we run two masking experiments. For the first experiment, we mask all the content words in each reference and ASR transcript and then calculate the cpWER on the partially masked transcripts with only the style words remaining. Following the PertLE Grande schema in~\citet{wang2023}, we run the Stanford NLP Group's Stanza tokenizer and part-of-speech (POS) tagger~\citep{qi2020stanza} on each transcript, replacing all nouns, proper nouns, main verbs, adjectives, and adverbs (in other words, maximally masking content words), according to their Universal POS tag, with SBERT's masking token, \verb!<mask>!. We then calculate the cpWER using the same procedure as before, comparing the masked Fisher reference transcripts to the masked ASR transcripts for each ASR system. For the second experiment, we do the reverse, masking all style words (i.e.~all words that are \emph{not} nouns, proper nouns, main verbs, adjectives, or adverbs) and calculate the cpWER on the partially masked transcripts with only the content words remaining.

\begin{table}[t]
\centering
\scalebox{0.87}{
\begin{tabular}{l|c|c|c} 
\bf ASR system & \bf Unmasked & \bf Mask$_C$ & \bf Mask$_{S}$ \\
\toprule
Gigaspeech & 14.0\% & 12.6\% & 10.8\% \\ 
AssemblyAI & 18.9\% & 18.2\% & 15.6\% \\
Switchboard & 20.8\% & 17.7\% & 14.9\% \\
Whisper & 26.4\% & 24.2\% & 20.8\% \\
TED-LIUM3 & 32.1\% & 28.0\% & 20.9\% \\
\bottomrule
\end{tabular}}
\caption{\small cpWER on unmasked, masked content words (Mask$_C$), and masked style words (Mask$_{S}$) for each ASR system compared to the reference Fisher gold standard test set (or the corresponding masked Fisher version). Significantly lower cpWER on Mask$_{S}$ than on Mask$_{C}$ suggests more errors occur on style words.}
\label{table:wer}
\vspace{-10pt}
\end{table}

The cpWER results for masking the content words (Mask$_C$) and masking the style words (Mask$_S$) are in~\autoref{table:wer}. Across all ASR systems, the cpWER decreases significantly more on the Mask$_S$ transcripts than on the Mask$_C$ transcripts. This difference could be because a notable portion of the total error comes from style words. Approximately 60\% of the total words per dataset are content words and approximately 40\% are style words, so masking style words leaves a majority of the dataset unmasked. One might expect more unmasked words to mean there are more words available for error and thus the cpWER would be higher on Mask$_S$; however, the opposite occurs, in which masking style words produces a \emph{lower} cpWER. Therefore, the errors are likely occurring on style words and masking them reduces the error rate. Moreover, the percentage of content and style words stays constant across all ASR transcriptions despite increases in the cpWER, indicating that the error rates are not simply due to the underlying distribution of content and style words in the data.

Admittedly, masking in this way is an imperfect measure because tokenization and POS tagging introduce additional errors. For instance, contractions like \emph{i'm} are tokenized as \emph{i 'm}, which when masking style words results in two tokens, \verb!<mask>! \verb!<mask>!, rather than one.\footnote{To help reduce token increases, when contractions involve only one masked component, we recombine both into one token (e.g.~\emph{wanna} $\rightarrow$ $<mask>$na).} Whether \emph{i'm} is deleted in the ASR transcript (but exists in the Fisher transcript) or is substituted or inserted in the ASR transcript (and does not exist in the Fisher transcript), splitting the contraction into two masked tokens contributes a separate masking error to the WER. 

Despite this possibility, a closer examination of the errors made by each ASR system reveals that nearly all of the most common errors are on style words, particularly due to transcription conventions (\emph{yeah} $>$ \emph{yes}, \emph{cause} $>$ \emph{because}, \emph{gonna} $>$ \emph{going to}), filler words and repeated restart deletions (e.g.~\emph{th- th- that's} vs.~\emph{that's}), and different filler word spellings (e.g.~\emph{umm} vs.~\emph{um}). However, if these style-based transcriptions are in fact consistent within the ASR system(s), then this kind of error may not significantly impact attribution performance, which may help explain why models incorporating style, such as StyleD and LUAR, do not show degradation. SBERT's poor performance across all transcriptions, but notably well below chance, is most likely due to its use of content words, which do not serve as dependable indicators in this `hard' topic manipulation setting. To test how both these style word errors and the presence of content words might impact attribution performance, we run the out-of-the-box attribution experiment on transcripts with their content words masked.

\paragraph{Removing content improves performance for models that rely on content.}\label{sec:masking}
Using the same masking schema as before, we mask all content words in the test set verification trials based on their POS tag. Then, we use the out-of-the-box attribution models to create embeddings of these masked test trials and evaluate the cosine similarity between speaker embeddings in trials using an AUC score. 

\begin{table}[t]
\centering
\scalebox{0.9}{
\begin{tabular}{r|c|c} 
$\Delta$ \emph{masking} & \bf Gold & \bf Mean across ASRs\\
\toprule
\bf SBERT & 0.267 & 0.320 \\
\bf StyleD & 0.038 & \textminus0.013 \\
\bf LUAR & 0.076 & 0.038 \\
\bottomrule
\end{tabular}
}
\caption{\small Difference between masked and unmasked out-of-the-box attribution performance (AUC) on gold standard and all ASR transcripts (mean) for each attribution model (rows). SBERT shows significant improvement without content words to mislead it; the more style-based models show no or only slight improvement.}
\label{tab:maskdelta}
\vspace{-.5cm}
\end{table}

\hyperref[tab:maskdelta]{Table 4} shows the difference in attribution performance on the masked test trials versus on the unmasked test trials (i.e.\ the original results from~\autoref{exp:auto}) on the gold standard Fisher versus on all ASR systems (averaged). Performance for all models and ASR systems individually is shown in~\autoref{tab:mask} in~\autoref{sec:appendix}. As expected, SBERT shows substantial improvement on the masked data over the unmasked data on both Fisher and the ASR systems because masking helps SBERT ignore content words, which are not reliable indicators in topic control settings (i.e.\ trials composed of same speakers discussing different topics and different speakers discussing the same topic). LUAR, which incorporates both style and content information into its decision-making, also shows some improvement. The style-based model StyleD does not degrade significantly despite a significant portion of the data being masked ($\sim$60\%), which aligns with the findings in~\citet{wang2023}. 

\subsection{Where is the floor in performance?}\label{exp:highwer}
\begin{figure}
\begin{center}
\hspace*{-.5cm} 
\begin{tikzpicture}[scale=0.95] 
    \begin{axis}[
        ybar stacked,
        ymin=0, ymax=1,
        symbolic x coords={Gold, DEU, R, R$_{\overline{U}}$, R$_{U_{10}}$, R$_{T_{50},U_{10}}$},
        xtick=data,
        bar width=15pt,
        legend style={at={(0.5,0.98)}, anchor=north},
        legend columns=-1, % Puts all legend items in a single row
        xlabel style={font=\small},
        xticklabel style={font=\small},
        ylabel style={at={(axis description cs:0,1.03)}, anchor=south, rotate=-90}, % Moves y-label to the top ticklabel
        ytick={0, 0.2, 0.4, 0.6, 0.8, 1.0}, % Define y-ticks
        yticklabels={0, 0.2, 0.4, 0.6, 0.8, AUC}, % Replace 1 with "AUC"
        yticklabel style={font=\small},
        height=5.5cm,  % Set a specific height for the plot
        width=8.5cm  % Keep the width the same
    ]
        % Min values
        \addplot [fill=blue] coordinates {(Gold,0.456162) (DEU,0.653102) (R,0.586) (R$_{\overline{U}}$,0.566000) (R$_{U_{10}}$,0.487399) (R$_{T_{50},U_{10}}$,0.499)};
        % Difference between min and avg
        \addplot [fill=gray] coordinates {(Gold,0.187597) (DEU,0.0223148000000001) (R,0.0669152000000001) (R$_{\overline{U}}$,0.063634) (R$_{U_{10}}$,0.009314) (R$_{T_{50},U_{10}}$,0.00107160000000006)};
        % Difference between avg and max
        \addplot [fill=orange] coordinates {(Gold,0.067215) (DEU,0.0309651999999999) (R,0.0318818) (R$_{\overline{U}}$,0.021157) (R$_{U_{10}}$,0.018287) (R$_{T_{50},U_{10}}$,0.00106739999999994)};
        
        \legend{Min, Avg, Max}
    \end{axis}
    % Draw the horizontal line manually across the full width
        \begin{pgfinterruptboundingbox} % Prevents affecting figure size
            \draw[thick, black, dashed] ([yshift=1.95cm]current axis.south west) -- 
                                ([yshift=1.95cm]current axis.south east);
        \end{pgfinterruptboundingbox}
\end{tikzpicture} \vspace{-.4cm}
\caption{\small Minimum, average, and maximum (each color's height) attribution model AUC performance (y-axis) on gold standard, German ASR (DEU), R (replacing each token with the same word), R$_{\overline{U}}$ (R $+$ truncating utterances to the mean length for that speaker in that call), R$_{U_{10}}$ (R $+$ truncating all utterances to 10 tokens), and R$_{T_{50},U_{10}}$ (R $+$ truncating transcripts to 50 utterances and utterances to 10 tokens) test verification trials (x-axis). Attribution performance does not drop significantly unless all utterance lengths are equalized.}\label{fig:grmrndm}
\end{center}\vspace{-.4cm}
\end{figure}
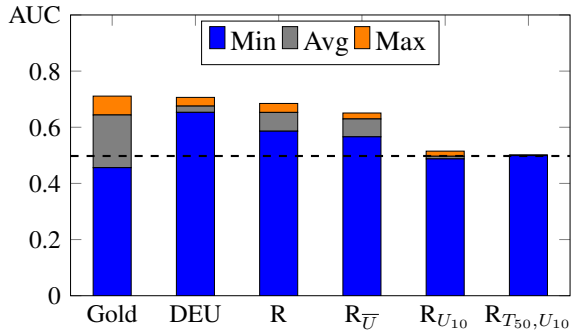

If attribution performance does not degrade at 32\% cpWER, then at what point \emph{will} performance degrade? To explore the impact on attribution performance when automatic transcription creates an even higher error rate, we first try transcribing the English audio using a different language ASR system. We choose German since German and English both use the Latin script and belong to the same language family, so transcriptions and potential phonological similarities can be easily compared. We choose a wav2vec 2.0 model fine-tuned on German using the CommonVoice dataset.\footnote{\href{https://huggingface.co/maxidl/wav2vec2-large-xlsr-german}{huggingface.co/maxidl/wav2vec2-large-xlsr-german}} The cpWER for this system is 90\%, thus significantly higher than those for the other ASR systems. Despite the transcripts containing mostly nonsensical German(-like) words, there are clear phonological similarities with some of the corresponding English words in the audio, such as the following transcriptions for the example in~\autoref{tab:transcr}: \emph{extremie} (extremely), \emph{agrie} (agree), \emph{shope aup} (soap operas). Therefore, there may still be revealing signal from each speaker's speech, which we in fact see in the performance in~\autoref{fig:grmrndm} (DEU). (\hyperref[tab:fullhighwer]{Table 10} in~\autoref{sec:appendix} provides the breakdown in AUC performance for each model.) The minimum model performance, SBERT, has increased significantly; the other models have roughly stayed the same or only slightly decreased, thus raising the overall average.\footnote{Although most of LUAR's Reddit training data was in English, there was a small portion in German, which could also contribute to LUAR's high performance on the German transcriptions. This SBERT model also saw a portion of Reddit during training so could have seen some German as well.} 

In previous work,~\citet{aggazzotti2024} found that SBERT can use noun overlap as a feature when making attribution decisions. We check to see if something similar might be happening, especially for the German ASR transcriptions, by measuring the number of shared unigrams and bigrams in positive trials versus in negative trials. In other words, even if each positive trial contains the same speaker discussing different topics, does that speaker still say particular words across topics more than different speakers discussing the same topic?

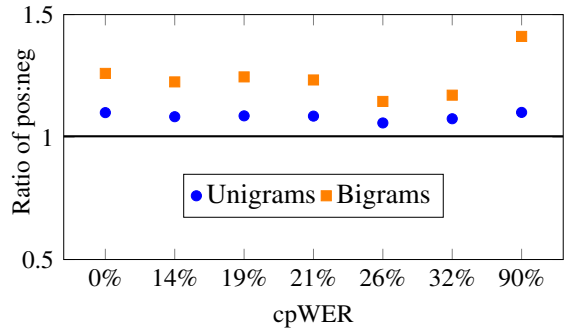
\begin{figure}[t]
\begin{center}
\hspace*{-.8cm} 
\vspace{-3mm}
\begin{center}
    \begin{tikzpicture}[scale=0.95] 
        \begin{axis}[
            symbolic x coords={0\%, 14\%, 19\%, 21\%, 26\%, 32\%, 90\%},
            xlabel={cpWER},
            xlabel style={font=\small},
            xticklabel style={font=\small},
            ylabel={Ratio of pos:neg},
            ylabel style={at={(axis description cs:-0.08,0.9)}, anchor=east},
            ymin=0.5, ymax=1.5,
            ytick={0.5, 1, 1.5},
            yticklabels={0.5, 1, 1.5},
            ylabel style={font=\small},
            yticklabel style={font=\small},
            legend style={at={(0.5,0.35)}, anchor=north},
            legend columns=-1, % Puts all legend items in a single row
            height=5cm,  % Set a specific height for the plot
            width=8.5cm  % Keep the width the same
        ]
            \addplot[only marks, mark=*, color=blue] coordinates {
                (0\%,1.099742) (14\%,1.083083) (19\%,1.086561) (21\%,1.085507) (26\%,1.057613) (32\%,1.074757) (90\%,1.100643) 
            };
            \addplot[only marks, mark=square*, color=orange] coordinates {
                (0\%,1.259898) (14\%,1.225195) (19\%,1.24561) (21\%,1.233086) (26\%,1.145437) (32\%,1.171046) (90\%,1.41088) 
            };
            \legend{Unigrams, Bigrams}
        \end{axis}
        \begin{pgfinterruptboundingbox} % Prevents affecting figure size
            \draw[thick, black] ([yshift=1.72cm]current axis.south west) -- 
                                ([yshift=1.72cm]current axis.south east);
        \end{pgfinterruptboundingbox}
    \end{tikzpicture}
\end{center}\vspace{-.5cm}
\caption{\small Ratio of unigram/bigram overlap in positive to negative trials (y-axis) across all ASR systems depicted by their cpWER (x-axis). All values are $>1$, indicating more overlap in positive trials, especially for German ASR (90\%) bigrams, helping explain why attribution performance does not significantly degrade. 
}\label{fig:overlap}
\end{center}\vspace{-.4cm}
\end{figure}

\paragraph{Word overlap can be a useful feature.} 
\autoref{fig:overlap} shows the ratio of unigram and bigram overlap between speakers in positive to negative trials across all ASR systems' transcriptions. All unigram and bigram overlap differences between positive and negative trials are statistically significant ($p < 0.05$ using an independent t-test). All data points are greater than one, meaning there is more unigram and bigram overlap in positive trials than in negative trials across all transcriptions. Since in this `hard' setting the conversational topic is controlled to be different for positive trials and the same for negative trials, only seeing slightly more unigram overlap in positive trials could mean that the topic manipulation is partially counteracting the overlap arising from being the same speaker. Nonetheless, especially the bigram overlap shows that speakers do have intraspeaker consistency in what they say regardless of conversational topic (positive trials) that is greater than different speakers discussing the same topic (negative trials). In particular, for the German ASR system at 90\% cpWER, despite being mostly nonsensical German, there are consistencies in the transcription that help the attribution models distinguish speakers so performance does not significantly degrade. Once again, SBERT is most likely using overlap as a clue for determining whether the speakers are the same or not, thus accounting for its significant improvement.
  
Since we still have not achieved a degradation in performance down to chance, we further maximize error in the transcription by substituting the same randomly selected word for each token in every trial. This procedure removes all content but preserves the utterance and transcript lengths.\footnote{The random word, \emph{lucubratory}, was selected randomly from the NLTK `words' corpus~\citep{nltk2009}.} The cpWER of this ``Random'' transcription (R) is 96\% (there are still some characters in common between the reference Fisher trials and the repeated random word trials), but as shown by the R bar in~\autoref{fig:grmrndm}, attribution performance decreases from that on the German transcription yet still does not degrade significantly. 

\paragraph{Utterance length is a surprisingly useful feature on this dataset.}
The only information remaining after removing the content is the utterance length and transcript length (number of utterances for a speaker in a call), in other words how much a person speaks. Therefore, the models must be availing themselves of one or both of these features as an indicator of whether the speakers are the same or not.\footnote{Even PANgrams, the minimum R performer, shows above chance performance because the $n$-grams are taken from across all of a speaker's utterances in a call (the speaker's utterances are concatenated into one string for that call) and are TF-IDF-weighted, so each word is counted with respect to its transcript frequency and its frequency across the training corpus overall. Therefore, some utterance and transcript length information is implicitly incorporated.} First, as a sanity check, we uniformly truncate all utterance lengths to 10 and all transcript lengths to 50 (both arbitrary) across all speakers (still using the one repeated word) and find that all models perform at chance, as expected (R$_{T_{50},U_{10}}$ in~\autoref{fig:grmrndm}). 

To tease apart the influence of utterance length versus transcript length, we devise two experiments. In one, we find the mean utterance length for each speaker in a trial and truncate all utterances to that length, still with all the tokens replaced by the fixed random word. Performance (R$_{\overline{U}}$) decreases only slightly from Random (R), suggesting that, in the absence of other information, the average utterance length per speaker can be revealing of speaker identity. We also do this same experiment but truncate the transcript length to 50 and get nearly identical performance. In the other experiment, we truncate all utterance lengths for each speaker to 10 (to be consistent with R$_{T_{50},U_{10}}$) but do not alter the transcript length (R$_{U_{10}}$). Now, finally, we see a drop in performance to chance, indicating that utterance length, but not necessarily the transcript length, can be a useful feature for these models on the Fisher LDC dataset.\footnote{We also tried shuffling the utterance order in case speakers' speaking more in the beginning or end of a call influenced performance, but the results were very similar to Random.} 
 
Despite these findings, we warn that this apparently significant effect of utterance length could be, at least in part, an artifact of the Fisher LDC dataset. We performed the same Random experiment on the BBN encoding of the same trials and all models' performance drops closer to chance ($\sim$0.67 to $\sim$0.57). Recall that the BBN encoding separates a speaker's speech into separate utterances if the other speaker interjects with a backchannel, while the LDC encoding groups a speaker's contiguous utterances together and places the backchannel after. 

\begin{table}[t]
\centering
\scalebox{0.9}{
\begin{tabular}{l|c|c|c}
 \emph{Fisher} & \bf Base & \bf Hard & \bf Harder\\
\toprule
\bf mean utt length (pos) & 12.9 & 12.8 & 12.8 \\ 
\bf mean utt length (neg) & 12.3 & 12.4 & 12.5 \\ 
\bottomrule
\end{tabular}%
}
\caption{\small Mean utterance length for Fisher positive and negative trials (rows) across all three difficulties (columns). Utterance length stays fairly constant across difficulties suggesting that it does not depend on topic.}
\label{tab:lengths}
\vspace{-.3cm}
\end{table}

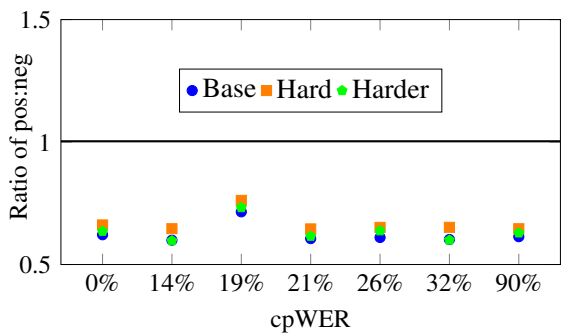
\begin{figure}[t]
\begin{center}
\hspace*{-.8cm} 
\vspace{-3mm}
\begin{center}
    \begin{tikzpicture}[scale=0.95] 
        \begin{axis}[
            symbolic x coords={0\%, 14\%, 19\%, 21\%, 26\%, 32\%, 90\%},
            xlabel={cpWER},
            xlabel style={font=\small},
            xticklabel style={font=\small},
            ylabel={Ratio of pos:neg},
            ylabel style={at={(axis description cs:-0.08,0.9)}, anchor=east},
            ymin=0.5, ymax=1.5,
            ytick={0.5, 1, 1.5},
            yticklabels={0.5, 1, 1.5},
            ylabel style={font=\small},
            yticklabel style={font=\small},
            legend style={at={(0.5,0.8)}, anchor=north},
            legend columns=-1, % Puts all legend items in a single row
            height=5cm,  % Set a specific height for the plot
            width=8.5cm  % Keep the width the same
        ]
            \addplot[only marks, mark=*, color=blue] coordinates { 
                (0\%,0.621482) (14\%,0.599337) (19\%,0.715051) (21\%,0.605715) (26\%,0.610337) (32\%,0.602295) (90\%,0.613506)  
            };
            \addplot[only marks, mark=square*, color=orange] coordinates {
            (0\%,0.661632) (14\%,0.646258) (19\%,0.761856) (21\%,0.645425) (26\%,0.651671) (32\%,0.651928) (90\%,0.646121) 
            };
            \addplot[only marks, mark=pentagon*, color=green] coordinates {
            (0\%,0.635868) (14\%,0.597291) (19\%,0.732473) (21\%,0.615494) (26\%,0.637695) (32\%,0.599954) (90\%,0.629046) 
            };
            \legend{Base, Hard, Harder}
        \end{axis}
        % Draw the horizontal line manually across the full width
        \begin{pgfinterruptboundingbox} % Prevents affecting figure size
            \draw[thick, black] ([yshift=1.72cm]current axis.south west) -- 
                                ([yshift=1.72cm]current axis.south east);
        \end{pgfinterruptboundingbox}
    \end{tikzpicture}
\end{center}\vspace{-.5cm}
\caption{\small Ratio of mean utterance length delta in positive to negative trials (y-axis) across all ASR systems depicted by their cpWER (x-axis) in all three difficulties. All values are $<1$, indicating a larger difference in utterance lengths in negative trials, and consistent across difficulty levels, suggesting that utterance length depends more on speaker than on topic. 
}\label{fig:lengths}
\end{center}\vspace{-.4cm}
\end{figure}

Since the Fisher calls involve two strangers discussing an assigned conversational topic, utterance length could also depend on the depth of the topic and how much the topic resonated with each speaker, which would help explain why performance on BBN still does not drop to chance. However, the average utterance length stays roughly the same in positive ($\sim$12.9) and negative ($\sim$12.4) trials across all three difficulty levels despite increased topic control, suggesting that utterance length does not depend on topic (see~\autoref{tab:lengths}). To further explore this finding, we test whether there is a significant difference between utterance length in positive versus negative trials by comparing the delta in utterance length for each transcript in a trial across all positive and negative trials for each ASR system's transcriptions in all three difficulties. As seen in~\autoref{fig:lengths}, utterance lengths are more different in negative trials than in positive trials in all settings ($p < 0.001$ using an independent t-test), and all difficulties are roughly the same, revealing that a speaker tends to have similar utterance lengths across calls, while different speakers vary more in their utterance length regardless of topic.

Another consideration is that models such as LUAR are conditioned on whatever information is available; thus, when only utterance length is available, LUAR uses that feature, but if there are other better features available, then it will make use of those as well (or instead). Therefore, the significance of utterance length might depend more on the presence of other features and less on the specific dataset, at least for some models. Future work should further explore the potential impact of utterance length on other speech transcript datasets. 

\subsection{Probing classifier}\label{exp:ft}
The attribution model representations considered thus far have not been specialized for the speech domain. Therefore, to assess the potential benefit of adapting them to this domain, we train a probing classifier that fits a multilayer perceptron (MLP) on top of each model's concatenated embeddings of each verification trial in the training set. We then evaluate the classifier on each model's concatenated embeddings of the test verification trials. We do this for each ASR transcription. Following~\citet{aggazzotti2024}, we do not fine-tune the whole model to avoid overfitting on the limited data, but fit a transformation of the fixed embedding from each attribution model. For PANgrams, though, we also follow the training protocol used by~\citet{pan2023} rather than that just described. Since the previous work found the setting in which the training and test data are from the same difficulty level to perform best, we only use this train-test difficulty match setting. Due to cost and time limitations, we only perform this experiment on Gigaspeech, Switchboard, and TED-LIUM3, but we expect similar results on the other two transcriptions, AssemblyAI and Whisper, since across all models and experiments performance is fairly consistent.

To more easily assess how well the probing classifier performs over the out-of-the-box results, we report the difference in AUC performance between the fine-tuned models and the out-of-the-box models. Since the deltas per ASR system are rather flat, we average the deltas across ASRs for each attribution model and report this averaged result in the first column of results in~\hyperref[tab:ftdeltas]{Table 6}; however, the delta of each individual attribution model's performance for each ASR system is shown in~\hyperref[tab:ft_o]{Table 11} in~\autoref{sec:appendix}. Positive AUC scores indicate that the fine-tuned models have improved attribution performance over the out-of-the-box models. 

\begin{table}[t]
\centering
\scalebox{0.9}{
\begin{tabular}{r|c|c}
\bf $\Delta$ & \emph{probing} & \emph{generalization}\\
\toprule
\bf PANgrams & \textminus0.104 & \textminus0.346 \\ 
\bf SBERT & \bf0.431 & \bf\textminus0.009 \\ 
\bf CISR & \textminus0.023 & \underline{\textminus0.013} \\ 
\bf StyleD & \textminus0.019 & \textminus0.041 \\ 
\bf LUAR & \underline{0.122} & \textminus0.101 \\ 
\bottomrule
\end{tabular}%
}
\caption{\small Mean difference between probing classifier and out-of-the-box performance (\emph{probing}) and train-test mismatch and match performance (\emph{generalization}) for each attribution model (rows). Largest improvements/least degradations per setting are bolded, second largest underlined. Training significantly helps SBERT and somewhat LUAR, but no model generalizes well from gold standard to automatic transcripts.}
\label{tab:ftdeltas}
\vspace{-.5cm}
\end{table}

\paragraph{Training only helps SBERT and LUAR}
SBERT shows the most improvement from training the probing classifier, and LUAR has the next highest. The remaining models and settings show a negative delta, meaning slightly worse performance from fine-tuning. Once again, SBERT's large improvement in performance from fine-tuning is most likely due to its availing itself of overlap between speakers in verification trials as a negative indicator of whether the speakers are the same.

\subsection{Generalization to ASR data}
\label{exp:mismatch}
In the previous probing experiment, each training set of verification trials matched the test set verification trials in regard to ASR system. This use case is helpful when there is enough transcribed data to both train and test on. However, in some real-world cases, the attribution model might have been trained on one kind of data or transcription, but must be evaluated on another. This experimental setup creates a train-test mismatch, but since we ideally would like a general purpose model, it is important to see how the models perform under such conditions. Relevant for this work is the case of training the models on the gold standard transcripts and evaluating them on each ASR transcript. Do models trained on human-transcribed data generalize to automatically-transcribed data? 

The second column in~\hyperref[tab:ftdeltas]{Table 6} shows the mean difference in performance between the train-test mismatch condition (training data is Fisher training verification trials and test data is each ASR's test verification trials) and the previous train-test match condition (training and test data for each ASR system are both from that ASR). The individual results per model are shown in~\hyperref[tab:ft_o]{Table 12} in~\autoref{sec:appendix}. The train-test mismatch task is slightly harder and none of the models perform as well, indicating they do not generalize well from gold standard to automatic transcripts. 

\section{Discussion}\label{sec:discussion}\paragraph{Summary of findings} Overall, our findings suggest that speaker attribution is surprisingly resilient to automatic transcription errors. In fact, our best result using out-of-the-box models on the `hard' difficulty level, $0.74$ AUC, is obtained using an ASR system with $21\%$ cpWER, compared to $0.71$ AUC using the reference transcripts. This is also true for models trained on transcripts, where the best result is $0.88$ AUC using Gigaspeech transcripts, which have $14\%$ cpWER. One possible explanation for these results is that the automatic transcripts are more faithful to the source, revealing more speaker-specific signal that can be exploited by attribution models. Our results also suggest that error consistency could play a role in the resiliency of speaker attribution to errors in transcription. We find that, in the absence of other information, the distribution of utterance lengths in a conversation can be a strong indicator of speaker identity, which establishes a lower bound on attribution performance even when the transcript itself is unreliable.

\paragraph{Broader impact} Our experiments paint an overall positive picture of the resiliency of speaker attribution to automatic transcription. However, for models trained on written text and applied out-of-the-box, their insensitivity to transcription errors may at least in part be explained by their inability to completely avail themselves of features of transcribed speech, and it is likely that models specialized to speech transcripts, and in particular specific ASR systems, may be more sensitive to changes in transcription at test time. Thus, our findings motivate further work that aims to learn specialized representations of speech transcripts for attribution purposes. 

\paragraph{Limitations and future work} 
In order to fairly compare transcriptions across ASR systems, for all systems except AssemblyAI (for which we transcribed by channel in the hopes of obtaining better transcriptions), we use Fisher's gold standard time segments for diarization. Although this utterance segmentation strategy can be fairly robust, incorrect segmentation is inevitable, especially when the speakers talk at the same time. Future work could try different methods of diarization, such as using pyannote.\footnote{\href{https://github.com/pyannote/pyannote-audio}{github.com/pyannote/pyannote-audio}} Further experiments with more verbatim ASR systems that capture more of a speaker's style and speaking patterns could help determine the impact of audio faithfulness on speaker attribution performance. One caveat to increased audio faithfulness is the scenario in which Speaker B clones Speaker A's voice and makes it say something Speaker B would say. The content of the speech might indicate Speaker B, but any transcription errors would likely indicate Speaker A. These conflicting signals might confuse the attribution system rather than improve performance, so in this case, a less faithful transcription might in fact be better. 

Pretraining a dedicated model on speech transcripts could better reveal the true capabilities of attribution models for speech transcripts, but would be difficult to do at scale since all of the speech would need to be transcribed using the same ASR system. Also, exploring the effect of utterance length in other speech transcript datasets could reveal how much of a confounding factor it is across settings. Finally, a separate interesting experiment would be to see whether the same attribution results hold for errors derived from translation, in other words, translated transcribed speech. 

\paragraph{Ethical considerations} We caution against the application of transcript-based speaker attribution methods in high-stakes settings, such as forensic cases, since they are not robust enough to potential spurious correlations. The best speaker attribution results on the `hard' difficulty level approach $0.90$ AUC by training the attribution models on speech transcripts. However, we caution that topic is controlled such that it correlates negatively with speaker identity in our selected trials. As a result, trained systems (notably SBERT, which performs worse than chance otherwise) can exploit this anti-correlation to perform well; therefore, out-of-the-box results, which are markedly lower (approximately $0.70$ AUC), may be more indicative of expected performance. Additionally, although the `base' difficulty level (without topic control, $0.84$ AUC) may seem more natural, the extent of topic shifts in the data may be difficult to determine, and thus the lower, more conservative out-of-the-box `hard' performance may be more realistic.

\section*{Acknowledgments}We thank the TACL reviewers and action editor for
their insightful comments. This research is supported in part by the Office of the Director of National Intelligence (ODNI), Intelligence Advanced Research Projects Activity (IARPA), via the HIATUS Program contract \#D2022-2205150003. The views and conclusions contained herein are those of the authors and should not be interpreted as necessarily representing the official policies, either expressed or implied, of ODNI, IARPA, or the U.S. Government. The U.S. Government is authorized to reproduce and distribute reprints for governmental purposes notwithstanding any copyright annotation therein.

\bibliography{references}
\bibliographystyle{acl_natbib}

\appendix
\section{Appendix}\label{sec:appendix}\begin{table*}
\centering
\begin{tabular}{@{\hskip 2pt}r@{\hskip 4pt}|cccccc} 
\multirow{2}{*}{AUC $\uparrow$} & \bf Gold & \bf Gigaspeech & \bf AssemblyAI & \bf Switchboard & \bf Whisper & \bf TED-LIUM3 \\
& \bf 0\% & \bf 14\% & \bf 19\% & \bf 21\% & \bf 26\% & \bf 32\% \\
\toprule
\bf PANgrams & \bf{0.805} & 0.783 & 0.783 & \underline{0.790} & 0.743 & 0.755 \\
\bf SBERT & \bf0.653 & 0.626 & 0.617 & \underline{0.640} & 0.592 & 0.626 \\
\bf CISR & 0.680 & \underline{0.694} & 0.661 & \bf0.699 & 0.676 & 0.691 \\ 
\bf StyleD & 0.698 & \underline{0.725} & 0.693 & \bf{0.739} & 0.707 & 0.702 \\
\bf LUAR & 0.803 & \bf{0.844} & 0.835 & \underline{0.843} & 0.802 & 0.820 \\
\bottomrule
\end{tabular}
\caption{\small Out-of-the-box AUC performance of each attribution model (rows) on \textbf{`base'} difficulty level gold standard Fisher and several automatic transcription test verification trials with various cpWERs (columns). Best transcription performance within each model (by row) is bolded and second best underlined, of which the differences are all statistically significant ($p < 0.001$ using a paired t-test). Despite considerable increases in cpWER, attribution performance surprisingly stays fairly constant, or even improves. The overall best performing transcription is Switchboard (21\% cpWER).}
\label{tab:fullboxbase}
\end{table*}

\begin{table*}[!]
\centering
\begin{tabular}{@{\hskip 2pt}r@{\hskip 4pt}|cccccc} 
\multirow{2}{*}{AUC $\uparrow$} & \bf Gold & \bf Gigaspeech & \bf AssemblyAI & \bf Switchboard & \bf Whisper & \bf TED-LIUM3 \\
& \bf 0\% & \bf 14\% & \bf 19\% & \bf 21\% & \bf 26\% & \bf 32\% \\
\toprule
\bf PANgrams & \bf{0.710} & 0.693 & 0.696 & \underline{0.701} & 0.639 & 0.653 \\
\bf SBERT & \bf0.456 & 0.394 & 0.317 & 0.410 & 0.315 & \underline{0.412} \\
\bf CISR & 0.664 & \underline{0.673} & 0.651 & \bf0.685 & 0.672 & \underline{0.673} \\ 
\bf StyleD & 0.678 & \underline{0.700} & 0.671 & \bf{0.717} & 0.693 & 0.686 \\
\bf LUAR & 0.711 & \underline{0.729} & 0.728 & \bf0.739 & 0.669 & 0.716 \\
\bottomrule
\end{tabular}
\caption{\small Out-of-the-box AUC performance of each attribution model (rows) on \textbf{`hard'} difficulty level gold standard Fisher and several automatic transcription test verification trials with various cpWERs (columns). Best transcription performance within each model (by row) is bolded and second best underlined, of which the differences are all statistically significant except ties ($p < 0.01$ using a paired t-test). Despite considerable increases in cpWER, attribution performance surprisingly stays fairly constant, or even improves. The overall best performing transcription is Switchboard (21\% cpWER).}
\label{tab:fullbox}
\end{table*}

\begin{table*}
\centering
\begin{tabular}{@{\hskip 2pt}r@{\hskip 4pt}|cccccc} 
\multirow{2}{*}{ $\Delta$ \emph{masking}} & \bf Gold & \bf Gigaspeech & \bf AssemblyAI & \bf Switchboard & \bf Whisper & \bf TED-LIUM3 \\
& \bf 0\% & \bf 14\% & \bf 19\% & \bf 21\% & \bf 26\% & \bf 32\% \\
\toprule
\bf PANgrams& \textminus0.004 & 0.010 & 0.013 & 0.002 & \bf0.022 & \underline{0.014} \\
\bf SBERT & 0.267 & 0.315 & \underline{0.338} & 0.300 & \bf0.358 & 0.288 \\
\bf CISR &  \bf 0.009 & \textminus0.017 & \underline{\textminus0.013} & \textminus0.015 & \textminus0.026 & \underline{\textminus0.013} \\
\bf StyleD & \bf0.038 & \textminus0.008 & \textminus0.007 & \textminus0.012 & \textminus0.031 & \underline{\textminus0.006} \\
\bf LUAR & \bf0.076 & 0.041 & 0.012 & 0.053 & \underline{0.055} & 0.028 \\
\bottomrule
\end{tabular}
\caption{\small Difference in attribution performance (AUC) of each out-of-the-box attribution model on test data with all content words masked and on unmasked test data across transcriptions with increasing error rates as measured by cpWER. Most improved performance from masking within each model (by row) is bolded and second best (or ties) underlined. }
\label{tab:mask}
\end{table*}

\begin{table*}
\centering
\begin{tabular}{@{\hskip 2pt}r@{\hskip 4pt}|cccccc} 
AUC $\uparrow$ & \bf Gold & \bf DEU & \bf Random & \bf R$_{\overline{U}}$ & \bf R$_{U_{10}}$ & \bf R$_{T_{50},U_{10}}$ \\
\toprule  
\bf PANgrams & \bf0.710 & \underline{0.670} & 0.586 & 0.566 & 0.515 & 0.499 \\
\bf SBERT & 0.456 & \underline{0.668} & \bf0.669 & 0.642 & 0.494 & 0.500 \\
\bf CISR & \bf0.664 & 0.653 & \underline{0.660} & 0.644 & 0.497 & 0.500 \\
\bf StyleD & \underline{0.678} & \bf0.680 & 0.665 & 0.646 & 0.487 & 0.500 \\
\bf LUAR & \bf0.711 & \underline{0.706} & 0.685 & 0.651 & 0.490 & 0.501 \\
\bottomrule
\end{tabular}
\caption{\small AUC performance of each attribution model (rows) on gold standard Fisher, German ASR (DEU; 90\% cpWER), R (replacing each token with the same word), R$_{\overline{U}}$ (R $+$ truncating utterances to the mean length for that speaker in that call), R$_{U_{10}}$ (R $+$ truncating all utterances to 10 tokens), and R$_{T_{50},U_{10}}$ (R $+$ truncating transcripts to 50 utterances and utterances to 10 tokens) test verification trials. Best transcription performance within each model (by row) is bolded and second best underlined. All differences among models within each transcription are statistically significant ($p < 0.01$ using a paired t-test). Performance for most models does not drop significantly until utterance length is removed as a feature (R$_{U_{10}}$ and R$_{T_{50},U_{10}}$). }
\label{tab:fullhighwer}
\end{table*}

\begin{table}
\centering
\scalebox{0.9}{
\begin{tabular}{@{\hskip 2pt}r@{\hskip 4pt}|c|ccc} 
\multirow{2}{*}{$\Delta$ \emph{probing}} & \bf Gold & \bf Giga & \bf Swb & \bf TED \\
& \bf 0\% & \bf 14\% & \bf 21\% & \bf 32\% \\
\toprule
\bf PANgrams& \underline{\textminus0.087} & \textminus0.118 & \bf\textminus0.077 & \textminus0.117 \\
\bf SBERT & 0.374 & \bf0.455 & \underline{0.425} & 0.414 \\
\bf CISR &  \textminus0.023 & \bf\textminus0.014 & \textminus0.037 & \underline{\textminus0.018} \\
\bf StyleD & \underline{\textminus0.004} & \bf\textminus0.003 & \textminus0.046 & \textminus0.008 \\
\bf LUAR & \bf0.161 & \underline{0.149} & 0.097 & 0.120 \\
\bottomrule
\end{tabular}}
\caption{\small The difference between probing classifier and out-of-the-box AUC performance (\emph{probing}) on each transcription (columns) for each attribution model (rows). Largest improvements or least degradations from training  per model (by row) are bolded and second largest are underlined. Training significantly helps SBERT and somewhat helps LUAR.}
\label{tab:ft_o}
\end{table}

\begin{table}
\centering
\scalebox{0.9}{
\begin{tabular}{@{\hskip 2pt}r@{\hskip 4pt}|ccc} 
\multirow{2}{*}{ $\Delta$ \emph{generalization}} & \bf Giga & \bf Swb & \bf TED \\
 & \bf 14\% & \bf 21\% & \bf 32\% \\
\toprule
\bf PANgrams& \bf0.122 & \textminus0.624 & \underline{\textminus0.536} \\
\bf SBERT & \textminus0.019 & \bf 0.000 & \underline{\textminus0.010} \\
\bf CISR &  \textminus0.020 & \bf\textminus0.007 & \underline{\textminus0.012} \\
\bf StyleD & \underline{\textminus0.040} & \textminus0.061 & \bf\textminus0.022 \\
\bf LUAR & \textminus0.118 & \bf\textminus0.071 & \underline{\textminus0.114} \\
\bottomrule
\end{tabular}}
\caption{\small The difference between train-test mismatch and match AUC performance (\emph{generalization}) on each transcription (columns) for each attribution model (rows). Largest improvements or least degradations per setting (by row) are bolded and second largest are underlined. Models do not generalize well from gold standard to automatic transcripts.}
\label{tab:mis_ft}
\end{table}

\end{document}